\documentclass[draft]{agujournal2019}
\usepackage{url} 
\usepackage{soul}
\usepackage{amsmath}

\draftfalse
\journalname{Geophysical Research Letters}
\begin{document}
%
%
\title{A physics-constrained emulator for time-discretised landslide runout modelling}

%
%
\authors{Ye Chen\affil{1,2}, Lorenzo Nava\affil{3,4,2,5}, Maximillian Van Wyk de Vries\affil{5,2,6}, Fawu Wang\affil{1}}

\affiliation{1}{Department of Geological and Hydraulic Engineering, Tongji University, Shanghai, China}
\affiliation{2}{Department of Earth Sciences, University of Cambridge, Cambridge, UK}
\affiliation{3}{Department of Geography, King's College London, London, UK}
\affiliation{4}{King's Institute of Artificial Intelligence, King's College London, London, UK}
\affiliation{5}{Department of Geography, University of Cambridge, Cambridge, UK}
\affiliation{6}{Scott Polar Research Institute, University of Cambridge, Cambridge, UK}

\correspondingauthor{Ye Chen}{tj\_chenye@tongji.edu.cn}

%
%
\begin{keypoints}
\item It is demonstrated that a neural network-based emulator can effectively predict landslide spatial-temporal evolution on unseen topography.
\item Incorporating physical constraints improves the physical consistency of emulator predictions.
\item The full landslide evolution process can be reproduced in a time-discretised manner through autoregressive application of the emulator.
\end{keypoints}

%
%
\begin{abstract}
Numerical simulations of landslides are commonly used to forecast impact areas. Full physics-based models provide accurate predictions, but their computational expense can preclude their use for large parameter sweeps. To achieve efficient modelling for such cases, we developed a physics-constrained neural network to emulate full landslide runout in a time-discretised manner. This emulator is trained on over 200,000 samples, each consisting of a pair of landslide thicknesses and velocities on a given DEM with a fixed time interval. To constrain the model's physical consistency, we define physics-based loss functions based on mass conservation and the system's kinetic energy loss. The model demonstrates high fidelity across the test dataset, achieving root mean square errors of approximately 0.06~m for thickness and 0.11~m/s for velocity in single-step predictions. In autoregressive rollout, these errors evolved modestly to about 0.21~m and 0.09~m/s after 110 iterations, while greatly improving computational efficiency by more than 50 times.
\end{abstract}

\section*{Plain Language Summary}
Landslides involve mass of material moving downslope and can cause catastrophic damage to society. Anticipating how a landslide travels in time and space is critical for hazard prevention. Researchers and hazard practitioners commonly use numerical models to solve equations describing landslide dynamics and to predict landslide evolution. However, in practice, running a large number of numerical simulations is computationally expensive and time-consuming. This is because the exact parameters of a landslide system are often uncertain, requiring many parallel simulations to identify the most likely outcomes. To improve the efficiency in landslide runout modelling, we trained an artificial neural network as a substitute for conventional numerical solvers. This model acts as a predictor, estimating how landslide material will evolve in the next time step based on its current state, including position and velocity. Trained on DEMs collected globally and on numerical simulations covering the full runout process from initiation to deposition, the model demonstrates robustness across a wide range of natural terrains. With this emulator, it is possible not only to predict single-step evolution but also to reproduce the entire runout process by applying the model recursively.

%
%
\section{Introduction}
Landslides are one of the most widespread natural hazards, and can cause devastating damage to human life and critical infrastructure \cite{morino_landslides_2022, sharma_systematic_2024}. Assessment of landslide hazards includes the assessment of runout, which represents the extent that landslide material travels from its source area to its final depositional position. Accurate prediction of landslide runout is important, as it delineates the potential impact area of an event and thereby provides critical information for risk assessment and effective disaster management \cite{hungr_model_1995, mcdougall_2014_2017}. In recent years, the increasing frequency of landslides that are triggered by extreme weather associated with global climate change \cite{gariano_landslides_2016, alcantara-ayala_landslides_2025}, has intensified the demand for efficient predictive tools. However, developing such models remains challenging, as it requires balancing computational efficiency with the level of physical fidelity and predictive accuracy needed for practical applications.

Numerical models serve as very valuable tools for quantitative landslide runout analysis \cite{hungr_two_2009}. By computationally solving governing equations of motion, and incorporating complex physical mechanisms such as frictional behaviour, material constitutive models, and multiphase coupling, these models can provide reliable motion details, regarding runout distance, peak velocity, motion duration \cite{crosta_numerical_2003, aaron_advancement_2017, chen_unified_2022}. However, this high level of predictive accuracy often comes at the expense of computational cost \cite{zhang_review_2024, zhao_artificial_2024}. While high-fidelity simulations are suitable for retrospective post-event analysis, their intensive resource requirements render them less practical for pre-event forecasting. This is particularly true for clustered landslide assessments or parameter calibration, where the necessity of running repetitive simulations with incremental parameter variations makes traditional numerical approaches computationally prohibitive.

In recent years, AI-based surrogate models, also known as emulators, have emerged as promising tools for the rapid approximation of complex physical systems. In general, surrogate models can achieve higher predictive accuracy than purely statistical models, while remaining more uncertain than fully physics-based numerical models. A wide range of advanced deep learning architectures, including convolutional neural networks, graph neural networks, and Fourier neural operators, are currently being explored for their potential to accelerate numerical simulations, with particularly notable success in fields such as fluid dynamics \cite{bender_feasibility_2019, li_fourier_2021, jiang_graph_2023, donnelly_physics-informed_2024}. Applications of surrogate model to landslide runout simulation are also under active development. Recent work has demonstrated the neural network's ability to map landslide evolution from $t_0$ to the final deposit on real world topographies \cite{nava_neural_2025}. However, this approach neglects intermediate states. This omission can limit their suitability for impact estimation in risk assessment, where temporal evolution is crucial. 

\begin{figure}
	\noindent\includegraphics[width=\textwidth]{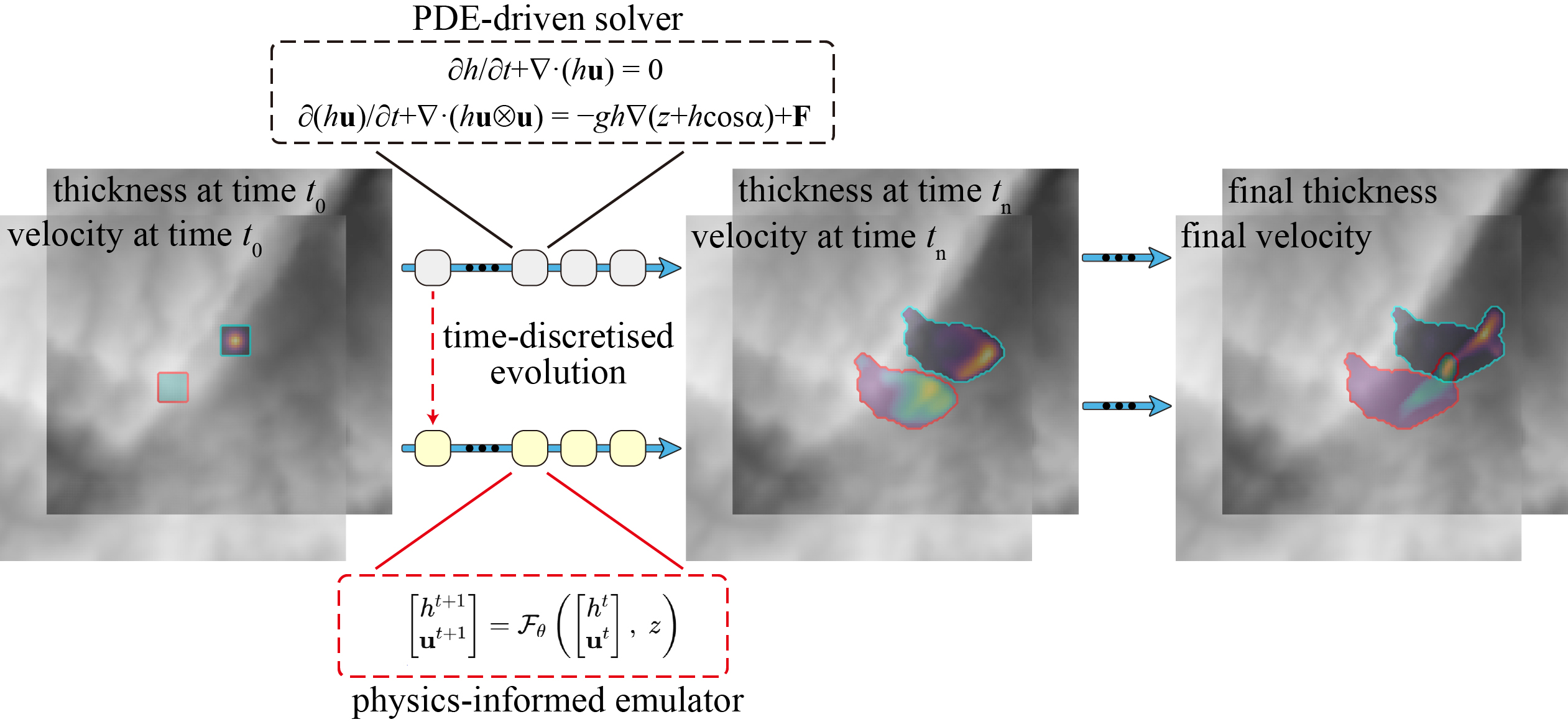}
    \caption{\textbf{Schematic framework of the physics-constrained emulator for time-discretised landslide runout modelling.} The emulator is trained to learn the mapping from landslide material states and terrain features from time \textit{t} to those at time \textit{t} + 1. It serves as a substitute for PDE-driven numerical solvers in the spatiotemporal simulation of landslide runout.}
    \label{fig:schematic_framework}
\end{figure}

Here we explore methods to overcome this limitation. This study develops and evaluates an artificial neural network based surrogate model, i.e. a time-discretised physics-constrained emulator designed for predicting the landslide runout spatiotemporally through complex mountainous terrain. The model is trained on more than 200,000 evolution pairs consisting of physical fields from consecutive time steps, generated by numerical simulations of depth-averaged flow model. The trained emulator predicts the propagation of landslide material based on its state at the previous time step, replacing the need for a PDE-driven solver (Fig. \ref{fig:schematic_framework}). When applied autoregressively, the emulator reproduces the full evolution of landslide runout. To ensure physical consistency and mitigate the error accumulation during the rollout predictions, the training was constrained by fundamental physical concepts via mass conservation and kinetic energy loss. The results demonstrate that this time-discretised emulator achieves high-fidelity performance in predicting entire landslide sequences while significantly reducing computational costs. Furthermore, the model architecture suggests substantial potential for incorporating additional physical mechanisms, facilitating its application in real-world disaster prevention.

\section{Methodologies}
The numerical model targeted for emulation is a two-dimensional depth-averaged flow model, which describes the evolution of landslide thickness and velocity magnitude. The training dataset was synthesised through a series of these numerical simulations. This model was selected because it captures essential kinetic features without the over-complication of excessive physical parameters, providing a clear baseline for evaluating the surrogate model's performance in this initial study. Meanwhile, the primary variables used in simulations, including topographic elevation, thickness, and velocity, are represented as 2D matrices. This data format is spatially analogous to multi-channel image formats, thereby enabling the integration of convolutional neural network architectures typically optimised for computer vision to solve these complex physical problems.

\subsection{Numerical model}
Depth-averaged flow models are widely adopted in established simulation frameworks for flow-like hazards. In this approach, the landslide mass is treated as a continuum governed by depth-integrated Navior-Stokes equations. Because of its flexible formulation, the depth-averaged framework can be readily adapted to describe different types of flow-like hazards by incorporating hazard-specifics kinetic and rheological characteristics. In this study, we basically follow the Volcflow model originally developed for simulating pyroclastic density currents \cite{kelfoun_numerical_2005}, and adapt it to landslide dynamics by representing the failed material as a viscous fluid with uniform material properties. The dynamics are governed by the conservation of mass and momentum, as expressed in the following integrated forms

\begin{equation}
    \frac{\partial h}{\partial t}
    + \frac{\partial (hu)}{\partial x}
    + \frac{\partial (hv)}{\partial y} = 0
    \label{eq:mass_conservation}
\end{equation}

\begin{equation}
    \frac{\partial (hu)}{\partial t}
    + \frac{\partial (hu^2)}{\partial x}
    + \frac{\partial (huv)}{\partial y}
    = gh \sin \alpha_x
    - \frac{1}{2} \frac{\partial}{\partial x} \left( g h^2 \cos \alpha_x \right)
    + F_x
    \label{eq: momentum_conservation_x}
\end{equation}

\begin{equation}
    \frac{\partial (hv)}{\partial t}
    + \frac{\partial (hvu)}{\partial x}
    + \frac{\partial (hv^2)}{\partial y}
    = gh \sin \alpha_y
    - \frac{1}{2} \frac{\partial}{\partial y} \left( g h^2 \cos \alpha_y \right)
    + F_y
    \label{eq: momentum_conservation_y}
\end{equation}

\noindent where \textit{h} denotes flow thickness, \textit{u} and \textit{v} represent velocity components in the Cartesian directions, $\alpha$ is the local ground slope, \textit{g} is the gravitational acceleration, and \textit{F} represents the momentum source term associated with basal friction. These governing equations are particularly effective when the horizontal spatial patterns are of greater significance than vertical velocity gradients, making it ideal for flow-like landslide runout analysis. Meanwhile, gravity and friction, only two dominate factors that cannot be omitted from landslide motion, comprise the main source terms in the current study. Other complex physical mechanisms of landslide motion have not been considered.

Basal friction is represented using the Voellmy friction model, a widely adopted approach originally developed for avalanche dynamics \cite{salm_short_2004}. In this study, a turbulent friction coefficient of  0.02~$\mathrm{m/s^2}$ was used. In this model, the coefficient determines the proportionality between the velocity-dependent basal shear stress and the square of the flow velocity. The rheology of the landslide material is described using a Bingham-type formulation, where the yield stress is controlled by a single rheological parameter, the cohesion. The resulting configuration allows for the generation of a wide range of flow mobilities while maintaining a minimal parameter set for the evaluation.

To numerically solve the depth-averaged model, we used a staggered finite volume method. The spatial computational domain was discretised into a regular structured grid of 256 $\times$ 256 cells with a constant spacing of \textit{H} = 30~m in both the \textit{x} and \textit{y} directions. Additional numerical constraints were implemented to further maintain computational stability and physical realism. A velocity threshold of 0.5~m/s was established as the stopping criterion for landslide motion, while the minimum flow thickness was set to $5 \times 10^{-4}$~m to avoid the negative influence of some extremely shallow flow layer. For temporal integration, the solver started with a time step of 0.25~s, which was adaptively adjusted according to the Courant-Friedrichs-Lewy (CFL) condition, with a minimum allowable step of 0.05~s.

\subsection{Global terrain-based runout dataset}
The fundamental objective of both the numerical and emulative models is to predict the evolution of landslide material across a given topography. Both frameworks require a digital elevation model (DEM) and an initial material thickness map as inputs to initiate a prediction. While the numerical simulator can adaptively process any physically plausible terrain by solving governing equations, the transferability of the emulator depends on a training dataset that encompasses a wide array of representative topographic features. Since the neural network aims to learn the underlying physics of gravity-driven motion over a DEM, a diverse selection of terrains is essential to ensure the model generalise effectively to global landscapes. 

Hence, we selected 19 mountainous regions worldwide, including Alaska, the Andes, the Alps, the Himalayas, and the Scandinavian Mountains, etc. Within each region, numerous non-overlapping DEM chips were extracted from the Copernicus Global Digital Elevation Model. Each chip consists of a 256 $\times$ 256 pixel matrix with a 30~m resolution, maintaining consistency with the cell structure of the numerical simulations. In total, 12,214 DEM maps were collected. The dataset features elevations ranging from -7.67~m to 8043.81~m, slope angles from 0.00° to 86.97°, and surface gradients from 0.00 to 18.88. 

Numerical simulations were performed on each of the collected DEMs. All simulations started with an identical thickness map. Although it neglects the varied boundary and thickness of source areas in real landslide cases, this unified initial condition helps us focus on discussion about the emulator's ability to learn terrain-dependent motion features preliminarily. This initial mass was configured as a central dome generated by a Gaussian distribution, featuring an 11 $\times$ 11 pixel footprint and a standard deviation ($\sigma$) of 0.2. With a total landslide volume set at $1\times10^6$~$\mathrm{m^3}$, the resulting maximum thickness was 37.4~m. The physical properties of the material were also held constant, with a density of 2,200~$\text{kg/m}^3$ and a cohesion of 10~kPa. All the simulations were recorded at a fixed temporal resolution of 1~s.

The final synthetic dataset is structured as a set of landslide evolution pairs $(t_n, t_{n+1})$. Each pair consists of thickness and velocity maps at time $t$ as input, with the corresponding fields at $t + \Delta t$ as output. Besides these dynamic fields, static topographic features are also incorporated into the evolution pairs. These static features include six tensors derived from the DEM: elevation, slope, aspect, curvature, and Cartesian coordinates (\textit{x}, \textit{y}), which can help the model to resolve the underlying terrain geometry and spatial context. Together with two dynamic variables, the final model input consists of an eight-layer stacked tensor. Since the emulator is formulated as a Markovian model, relying only on the current state to predict the next step, we randomly sampled 30\% of the available state-transition pairs from each simulation sequence, resulting in a final dataset of 228,662 samples.

\subsection{Emulative model trained by using PINNs}
We adopt an attention-based residual U-Net architecture, following \cite{nava_neural_2025}, where it was shown to perform effectively for landslide runout predictions. This architecture preserves spatial structure while enabling efficient feature extraction across scales. To improve stability and mitigate error propagation during the autoregressive process, physical concepts governing the system (i.e mass conservation and kinetic energy loss) are incorporated into the loss functions as physics-based constraints.

In runout prediction, physical fields such as flow thickness and velocity are spatially localised, as they exist only in regions where landslide material is present. Consequently, the data matrices contain predominantly zero values outside the landslide boundary, which occupy a much larger area than regions with meaningful values. To reduce the influence of these zero-dominated regions during training, a mask based on the ground-truth flow region is applied, and the loss functions are computed only within the masked area. The masked area is defined as cells with positive values (denormalised value $>$ 0.001) in each map.

Equation \ref{eq: loss_mass} defines the incorporation of the mass conservation principle into the loss function. In this formulation, $M_\text{pred}$ is the total mass of material over the predicted map, whereas $M_\text{gt}$ represents the corresponding mass over the ground-truth map. This loss is implemented by summing the flow thickness over the masked areas in the domain and comparing the resulting total volume (which represents mass, given the constant density in this study) with the ground-truth value. In this way, even if local errors exist, the global mass of the system is constrained to be conserved. The loss penalises predictions that artificially create or remove material.

\begin{equation}
    \mathcal{L}_{\mathrm{mass}}
    =
    \frac{
        \big| 
            M_{\mathrm{pred}}
            -
            M_{\mathrm{gt}}
        \big|
    }{
        \big| M_{\mathrm{gt}} \big|
    }
\label{eq: loss_mass}
\end{equation}

To constrain the velocity predictions, an additional physics-based loss is introduced. Although momentum or kinetic energy is not conserved throughout the landslide process - since the motion initiates from rest, undergoes acceleration and deceleration, and eventually returns to rest - it can still serve as a useful constraint in a manner analogous to the loss of mass, as shown in equation \ref{eq: loss_ke}. Specifically, the total kinetic energy of the system is computed at each sample and compared between the predicted and ground-truth maps to construct this loss. Here, $K_\text{pred}$ is the total kinetic energy of the predicted field, and $K_\text{gt}$ represents the corresponding value for ground-truth field.

\begin{equation}
    \mathcal{L}_{\mathrm{ke}}
    =
    \frac{
        \big|
            K_{\mathrm{pred}}
            -
            K_{\mathrm{gt}}
        \big|
    }{
        \big| K_{\mathrm{gt}} \big|
    }
\label{eq: loss_ke}
\end{equation}

Apart from the two physics-based losses above, a basic data loss using the Mean Absolute Error (MAE) $\mathcal{L}_{\mathrm{total}}$ is adopted, together with two auxiliary losses: background loss and smoothness loss. The background loss (equation \ref{eq: loss_bg}), operates exclusively on background regions where the ground-truth thickness is zero, penalising any non-zero predictions in these areas. In this formulation, $w^{\mathrm{bg}}_{ij}$ is set to 1.0 in background regions and 0 within the flow region. In addition, a smoothness loss is introduced (equation \ref{eq: loss_tv}). This term corresponds to the total variation loss, calculated as the sum of absolute differences between neighbouring pixels across the spatial domain, where \textit{H} and \textit{W} represent the height and width of the map. The smoothness loss acts to reduce noise and oscillatory artifacts in the predicted fields.

\begin{equation}
    \mathcal{L}_{\mathrm{bg}}
    =
    \frac{
    \displaystyle
    \sum_{i,j}
    \Big[
    w^{\mathrm{bg}}_{ij} \, \big| y_{ij} - \hat{y}_{ij} \big|
    \Big]
    }{
    \displaystyle
    \sum_{i,j} w^{\mathrm{bg}}_{ij}
    }
\label{eq: loss_bg}
\end{equation}

\begin{equation}
    \mathcal{L}_{\mathrm{tv}}
    =
    \frac{1}{HW}
    \left[
    \sum_{i=1}^{H-1}\sum_{j=1}^{W}
        \big| x_{i+1,j} - x_{i,j} \big|
    \;+\;
    \sum_{i=1}^{H}\sum_{j=1}^{W-1}
        \big| x_{i,j+1} - x_{i,j} \big|
    \right]
\label{eq: loss_tv}
\end{equation}

We divide the dataset into three subsets for training, validating, and testing, in proportions of 70\%, 20\%, and 10\%, respectively. During training of the time-discretised emulator, all the aforementioned loss terms are combined into a total loss function, as defined in equation \ref{eq: loss_total}. The hyperparameters, including the four loss weights and the threshold used for map masking, are optimised by searching over reasonable value ranges. This search is conducted over 80 training trials using a subset of the full dataset. The subset is constructed by selecting 10\% pairs from simulation sequences instead of 30\%, which preserves the representativeness of the training data while reducing computational cost. The optimised values are \(\lambda_{\mathrm{bg}} = 0.05\), \(\lambda_{\mathrm{tv}} = 0.01\), \(\lambda_{\mathrm{mass}} = 0.005\), and \(\lambda_{\mathrm{ke}} = 0.005\).

\begin{equation}
    \mathcal{L}_{\mathrm{total}}
    =
    \mathcal{L}_{\mathrm{m}}
    + 
    \lambda_{\mathrm{bg}}\,
    \mathcal{L}_{\mathrm{bg}}
    +
    \lambda_{\mathrm{tv}}\,
    \mathcal{L}_{\mathrm{tv}}
    +
    \lambda_{\mathrm{mass}}\,
    \mathcal{L}_{\mathrm{mass}}
    +
    \lambda_{\mathrm{ke}}\,
    \mathcal{L}_{\mathrm{ke}}
\label{eq: loss_total}
\end{equation}

\section{Results}
In this section, we evaluate the fidelity and computational performance of the emulated solutions relative to the numerically simulated results. Experiments conducted using the trained emulator, both with and without physics-based loss terms, are compared in terms of single-step predictions across the entire test dataset and autoregressive predictions over a randomly selected unseen DEM. This comparison enables an assessment of the model’s transferability, as well as the extent to which nonphysical deviations in the predictions can be mitigated through the incorporation of physical constraints.

\subsection{Fidelity of the emulator on time-discretised predictions}
The emulator performance was evaluated using four complementary metrics. Intersection over Union (IoU) assessed the accuracy of the predicted flow extent. Root Mean Square Error (RMSE) quantified pointwise prediction errors in physical units. The Pearson correlation coefficient measured spatial agreement between predicted and reference fields, capturing potential spatial misalignment not reflected by RMSE. Finally, global physical consistency was evaluated using total volume and total kinetic energy. Models trained with and without physics-based constraints were compared.

The models were evaluated across the entire test dataset, and the corresponding metric values are listed in Table \ref{table: evaluation metrics}. The results show that the physics-constrained emulator attains IoU scores of 0.989 for thickness and 0.904 for velocity, with corresponding RMSE values of 0.061~m for thickness and 0.110~m/s for velocity, and Pearson correlation coefficients of 0.999 and 0.993, respectively. The incorporation of physical constraints does not appear to significantly improve these standard performance metrics. In contrast, physically meaningful features, such as the total volume and maximum velocity, are much better constrained through physics-based loss functions. Specifically, the shifts in total volume are reduced from -2130~$m^3$ to 36~$m^3$, and the shifts in maximum velocity from -0.043~$m/s$ to -0.025~$m/s$. While these shifts in physical quantities may seem minor for individual predictions, they can accumulate during sequential rollout predictions, potentially leading to significant errors.

\begin{table}
    \caption{\textbf{Evaluation metrics on the emulators trained with and without physical constraints.}}
    \centering
    \begin{tabular}{l c c}
    \hline
     evaluation metrics & data-driven emulator & physics-constrained emulator \\
    \hline
      IoU (\textit{h})  & 0.965 ± 0.021 & 0.989 ± 0.009 \\
      IoU (\textit{v})  & 0.867 ± 0.091 & 0.904 ± 0.078 \\
      RMSE (\textit{h}) & 0.049 ± 0.076 & 0.061 ± 0.062 \\
      RMSE (\textit{v}) & 0.012 ± 0.031 & 0.110 ± 0.035 \\
      Pearson correlation coefficient (\textit{h}) & 0.999 ± 0.000 & 0.999 ± 0.000 \\
      Pearson correlation coefficient (\textit{v}) & 0.994 ± 0.052 & 0.993 ± 0.015 \\
      Shifts in total volume (\textit{h}) & -2130 ± 4204 & 36 ± 2120 \\
      Shifts in maximum velocity (\textit{v}) & -0.043 ± 0.538 & -0.025 ± 0.131\\
    \hline
    \end{tabular}
    \label{table: evaluation metrics}
\end{table}

\begin{figure}
	\centering
	\noindent\includegraphics[width=0.6\textwidth]{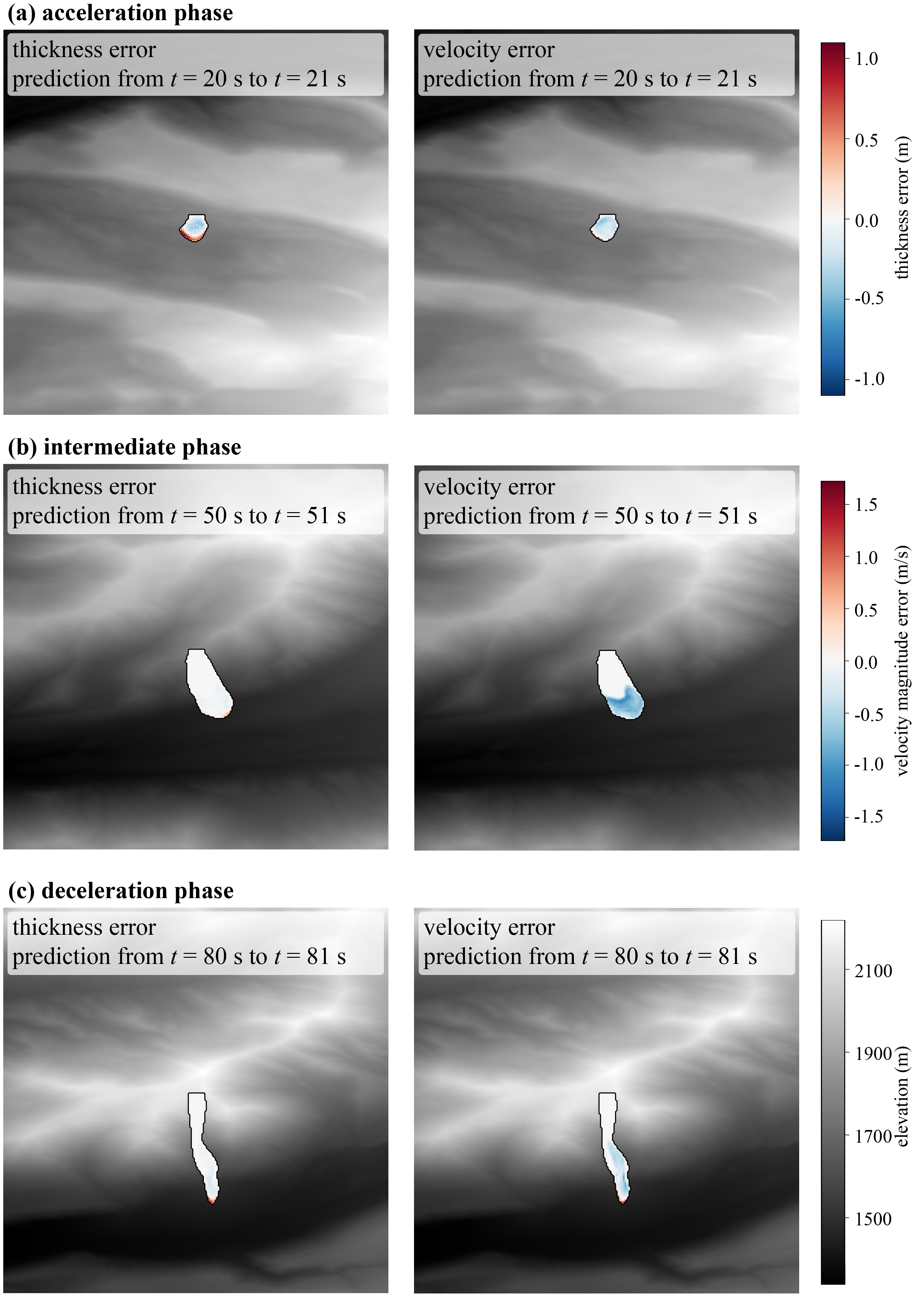}
	\caption{\textbf{Model performance for single step predictions on three unseen DEMs.} 
    \textbf{(a)} The thickness and velocity error maps tested in an acceleration phase of a landslide motion from time 20~s to 21~s. \textbf{(b)} The thickness and velocity error maps tested in an intermediate phase of a landslide motion from time 50~s to 51~s. \textbf{(c)} The thickness and velocity error maps tested in an deceleration phase of a landslide motion from time 80~s to 81~s.}
    \label{fig: one_step predictions}
\end{figure}

To visualise prediction errors, three DEMs and related runout simulations were randomly selected from the test dataset to evaluate the model’s performance at different phases of the runout: the acceleration phase, intermediate phase, and deceleration phase. Figure \ref{fig: one_step predictions} presents the corresponding error maps for single-step predictions in these phases. In these maps, the black boundary delineates the ground-truth landslide extent, while positive and negative prediction biases are shown in red and blue, respectively. Overall, the predictions exhibit high visual fidelity in terms of spatial extent, with IoU scores of 0.976, 0.989, and 0.989 for the thickness maps. The error patterns show that biases predominantly occur in regions with high flow velocities and are mostly negative. During the acceleration phase, biases are distributed relatively uniformly across the landslide area, as most material is undergoing rapid motion. In contrast, during the intermediate and deceleration phases, the largest biases are concentrated near the front regions within the landslide boundaries, with errors decreasing in the deceleration phase. This behaviour is physically interpretable, as regions with higher velocities correspond to more rapid changes in material motion, which pose greater challenges for accurate prediction. Across these test cases, the maximum bias is approximately 1~m in denormalised thickness and -1.5~m/s in denormalised velocity. Given the spatial resolution of 30~m, these errors are relatively small and considered acceptable for predictive modelling purposes.

\subsection{Fidelity of the emulator on autoregressive predictions}
\begin{figure}
	\centering
	\noindent\includegraphics[width=\textwidth]{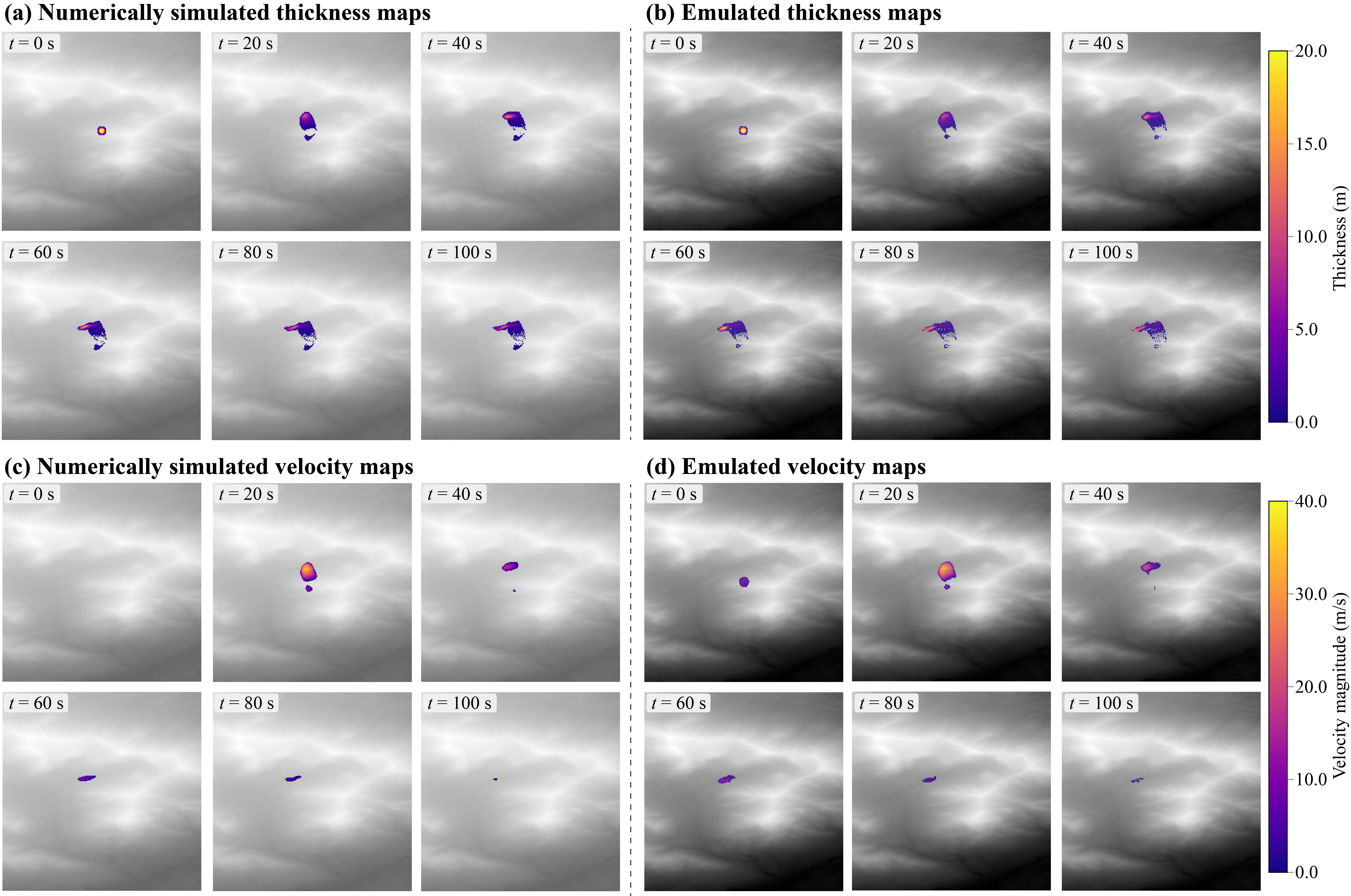}
	\caption{\textbf{Comparison between the simulated and emulated full sequence on an unseen DEM.}
		  \textbf{(a)} The thickness maps show the numerically simulated results with an time interval of 20~s. \textbf{(b)} The corresponding emulated thickness maps predicted on the same DEM. \textbf{(c)} The corresponding velocity magnitude maps produced by the numerical simulation. \textbf{(d)} The corresponding velocity magnitude maps predicted by the emulator.}
    \label{fig: test_dataset}
\end{figure}

To predict the complete temporal evolution of a landslide runout, the time-discretised emulator can be applied autoregressively in a rollout procedure. The emulator's predicted outputs are directly fed back as inputs for subsequent predictions. We terminate autoregressive rollout once the landslide mass ceases motion. Fig. \ref{fig: test_dataset} illustrates a predicted runout sequence over an unseen DEM in the Alaska region, emulated for 110 seconds (emulated steps) until the flow comes to rest. Although the predicted results do not perfectly replicate the numerical simulations, the final runout distance, movement trajectory, and overall boundary geometry are reproduced with high accuracy. 

To quantitatively assess prediction accuracy, the temporal evolution of three metrics is evaluated: the IoU of landslide material (based on the thickness map), and the RMSE of the predicted thickness and velocity maps. According to the statistics of the IoU over the full prediction sequence, the accuracy of landslide boundary prediction gradually decreases during autoregressive rollout, but stabilises after 30~s at a value of around 0.94. Meanwhile, the thickness error starts at approximately 0.08~m at the first prediction step, increases to 0.20~m after 60~s, and then plateaus. In contrast, the velocity error peaks at around 12~s, when the landslide moves most rapidly, and subsequently decreases to a level even lower than that of the initial prediction. These results demonstrate that the autoregressive predictions produced by the time-discretised emulator remain stable and achieve satisfactory accuracy over the full sequence.

\subsection{Physical consistency of the emulator}

\begin{figure}
	\centering
	\noindent\includegraphics[width=0.6\textwidth]{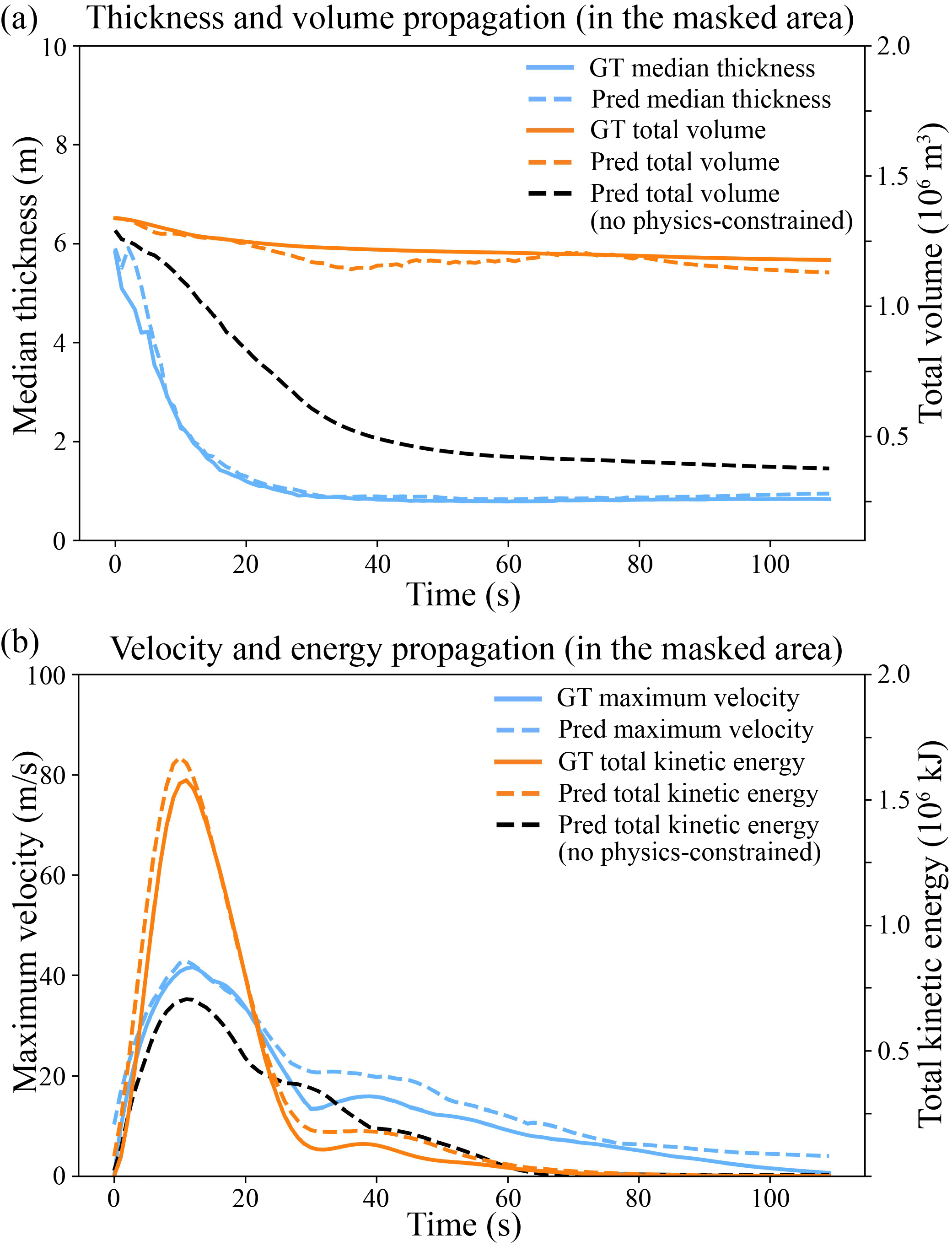}
	\caption{\textbf{Analysis on the physical consistency between the results provided via numerical simulator and emulator.}
		  \textbf{(a)} The evolution of mass features including median thickness and sum of thickness during the rollout prediction. \textbf{(b)} The evolution of kinetic features including maximum velocity magnitude and sum of kinetic energy during the rollout prediction.}
	\label{fig:physical_consistency}
\end{figure}

Statistical metrics alone are insufficient to assess the validity of predictions for physical systems, as accurate predictions may still violate fundamental physical constraints. For physics-constrained models, it is therefore necessary to evaluate both predictive accuracy and physical consistency, particularly with respect to mass conservation and the temporal evolution of velocity. To this end, we evaluate several physically meaningful quantities during rollout prediction: median thickness, total thickness (material volume), maximum velocity, and total kinetic energy. Fig.~\ref{fig:physical_consistency} shows their temporal evolution.

For thickness-related quantities, the predicted total thickness closely follows the ground-truth curve, with similarly good agreement in median thickness. This indicates that the model preserves the overall volume of the sliding mass. A slight decrease in total thickness is observed in both predicted and reference solutions, due to the removal of very shallow layers for numerical stability. This behaviour is also captured by the emulator. For velocity-related quantities, both maximum velocity and total kinetic energy show good agreement with the numerical solution, indicating that the model captures the main phases of acceleration and deceleration. Before the peak (approximately 10~s), maximum velocity is predicted more accurately, whereas after approximately 60~s, total kinetic energy shows better agreement. In particular, the model predicts a small residual maximum velocity (around 4~m/s) at late times, while the kinetic energy decays more consistently.

Additionally, the black dashed curves in Fig.~\ref{fig:physical_consistency} illustrate the rollout performance of the emulator without physics-based loss constraints. Small biases observed in single-step predictions become amplified during autoregressive rollout. In particular, the total landslide volume decreases substantially, with the final deposited volume reduced to nearly one-third of its initial value. The total kinetic energy of the system also decays prematurely, reaching zero approximately 20~s earlier than expected. These losses arise from persistent negative biases at each time step, which accumulate over time.

More importantly, the emulator spends approximately 10~ms per step and 1.1~s for a full 110~s rollout, compared to 500~ms per step and 16.7~s for the numerical simulator, corresponding to a speedup of about 50$\times$. Unlike numerical solvers, whose cost increases with model complexity, the emulator maintains a nearly constant inference time once trained. This makes it particularly suitable for large-scale or repeated simulations.

\section{Conclusions}
Artificial neural networks are increasingly being adopted as powerful surrogate models for complex numerical simulations. In this work, we establish the feasibility of using convolutional neural networks to construct a time-discretised surrogate model that emulates the landslide motion process by replacing a single iteration of the core numerical solver for the governing dynamic equations. The proposed emulator functions as a hybrid matrix-to-matrix regressor, learning the underlying dynamics that governs the propagation of failed material over complex terrain. It simultaneously predicts both thickness and velocity fields within a single inference step.

The model achieves acceptable accuracy in predicting landslide motion over durations exceeding 100~s, while offering a computational speedup of more than 50 times compared to the original numerical solver. To mitigate error accumulation inherent in autoregressive rollouts, additional physical constraints based on system mass conservation and kinetic energy loss are incorporated into the training process, resulting in substantially improved physical consistency of the predictions.

Despite these promising results, the current surrogate model is not yet sufficiently general for direct application to real-world scenarios. In particular, the training data encompass only a single flow mobility regime, and terrain-dependent frictional effects are not explicitly represented. Nevertheless, the framework can be readily extended by incorporating additional input information, such as material properties, initial source characteristics, spatially varying terrain friction, and other relevant physical mechanisms. Such extensions are expected to enable the development of a more robust and accurate surrogate model. The successful validation of this time-discretised predictor highlights its potential as a foundation for integrating increasingly complex and realistic physical processes into the prediction of evolving landslide motion systems.



%
%
\acknowledgments
This study is supported by the Key Project of the National Natural Science Foundation of China (Grant No. 42230715), and the National China Scholarship Council program (Project ID: 202406260107).

\section*{Open Research Section}
The source code and dataset used for training and testing the developed emulator are archived in Zenodo at https://doi.org/10.5281/zenodo.20151403 \cite{chen_2026_20151403}

\section*{Conflict of Interest}
The authors declare there are no conflicts of interest for this manuscript.

%
%

\bibliography{emulator}

\end{document}